# Character-Level Neural Translation for Multilingual Media Monitoring in the SUMMA Project

**Guntis Barzdins, Steve Renals, Didzis Gosko**
University of Latvia, University of Edinburgh, LETA
Riga 29 Rainis Blvd. IMCS UL, Edinburgh EH8 9AB, Riga 2 Marijas Str.
E-mail: guntis.barzdins@lu.lv, s.renals@ed.ac.uk, didzis.gosko@leta.lv

**Abstract**

The paper steps outside the comfort-zone of the traditional NLP tasks like automatic speech recognition (ASR) and machine translation (MT) to addresses two novel problems arising in the automated multilingual news monitoring: segmentation of the TV and radio program ASR transcripts into individual stories, and clustering of the individual stories coming from various sources and languages into storylines. Storyline clustering of stories covering the same events is an essential task for inquisitorial media monitoring. We address these two problems jointly by engaging the low-dimensional semantic representation capabilities of the sequence to sequence neural translation models. To enable joint multi-task learning for multilingual neural translation of morphologically rich languages we replace the attention mechanism with the sliding-window mechanism and operate the sequence to sequence neural translation model on the character-level rather than on the word-level. The story segmentation and storyline clustering problem is tackled by examining the low-dimensional vectors produced as a side-product of the neural translation process. The results of this paper describe a novel approach to the automatic story segmentation and storyline clustering problem.

## 1. The SUMMA Project Overview

Media monitoring enables the global news media to be viewed in terms of emerging trends, people in the news, and the evolution of storylines (Risen et al., 2013). The massive growth in the number of broadcast and Internet media channels requires innovative ways to cope with this increasing amount of data. It is the aim of SUMMA[1] project to significantly improve media monitoring by creating a platform to automate the analysis of media streams across many languages.

Within SUMMA project three European news broadcasters BBC, Deutche Welle, and Latvian news agency LETA are joining the forces with the University of Edinburgh, University College London, Swiss IDIAP Research Institute, Qatar Computing Research Institute, and Priberam Labs from Portugal to adapt the emerging big data neural deep learning NLP techniques to the needs of the international news monitoring industry.

BBC Monitoring undertakes one of the most advanced, comprehensive, and large scale media monitoring operations world-wide, providing news and information from media sources around the world. BBC monitoring journalists and analysts translate from over 30 languages into English, and follow approximately 13,500 sources, of which 1,500 are television broadcasters, 1,300 are radio, 3,700 are key news portals world-wide, 20 are commercial news feeds, and the rest are RSS feeds and selected Social Media sources. Monitoring journalists follow important stories and flag breaking news events as part of the routine monitoring.

The central idea behind SUMMA is to develop a scalable multilingual media monitoring platform (Fig.1) that combines the real-time media stream processing (speech recognition, machine translation, story clustering) with in-depth batch-oriented construction of a rich knowledge base of reported events and entities mentioned, enabling extractive summarization of the storylines in the news.

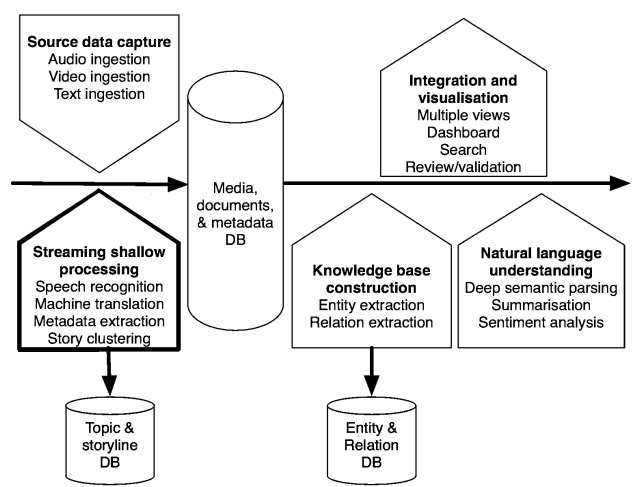

Figure 1: The components of the SUMMA project.

In this paper we focus only on the *streaming shallow processing* part of the SUMMA project (the dark block in Fig.1), where the recently developed neural machine translation techniques (Sutskerev, Vinyals & Le, 2014; Bahdanau, Cho & Bengio, 2014) enable radically new end-to-end approach to machine translation and clustering of the incoming news stories. The approach is informed by our previous work on machine learning (Barzdins, Paikens, Gosko, 2013), media monitoring (Barzdins et al.,2014), and character-level neural translation (Barzdins & Gosko, 2016).

---

[1] SUMMA (Scalable Understanding of Multilingual MediA) is project 688139 funded by the European Union H2020-ICT-16 BigData-research call. The project started in February 2016 and will last 3 years.

## 2. Multilingual Neural Translation

Automation of media monitoring tasks has been the focus of the number of earlier projects such as European Media Monitor (emm.newsbrief.eu), EventRegistry (eventregistry.org), xLike (xlike.org), Bison (bison-project.eu), NewsReader (newsreader-project.eu), MultiSensor (multisensorproject.eu), inEvent (invent-project.eu), and xLiMe project (xlime.eu). These predecessor projects are dominated by the paradigm of NLP pipelines [20] based on shallow machine learning.

| | language | audio ASR MT ETL SP | text MT ETL KB SP | social media MT ETL SP |
|---|---|---|---|---|
| good resources | Arabic | • • • • | • • • • | • • • |
| | English | • – • • | – • • • | – • • |
| | German | • • • • | • • • • | • • • |
| | Spanish | • • • • | • • • • | • • • |
| some resources | Portuguese | • • • ∘ | • • • ∘ | • • ∘ |
| | Russian | • • • ∘ | • • • ∘ | • • ∘ |
| | Farsi | • • • ∘ | • • • ∘ | • • ∘ |
| poor resrc. | Ukrainian | • • ∘ ∘ | • ∘ ∘ ∘ | • ∘ ∘ |
| | Latvian | • • ∘ ∘ | • ∘ ∘ ∘ | • ∘ ∘ |

**ASR**: Autom. Speech Recognition; **MT**: Machine Translation; **ETL**: Entity Tagging & Linking
**KB**: Knowledge Base Construction; **SP**: Semantic Parsing
• native processing  ∘ cross-lingual processing via MT into English  – not applicable

Table 1: Languages covered and their level of processing within SUMMA.

The key difference of the SUMMA project is that it has been incepted after the recent paradigm-shift (Manning, 2015) in the NLP community towards neural network inspired deep learning techniques such as end-to-end automatic speech recognition (Graves & Jaitly, 2014; Hannun et al., 2014; Amodei, 2015), end-to-end machine-translation (Sutskerev, Vinyals & Le, 2014; Bahdanau, Cho & Bengio, 2014; Luong et al., 2015), efficient distributed vectorspace word embeddings (Mikolov et al., 2013), image and video captioning (Xu et al., 2015; Venugopalan et al., 2015), unsupervised learning of document representations by autoencoders (Li, Luong & Jurafsky, 2015). These recent deep learning breakthroughs along with massively parallel GPU computing allow addressing the media monitoring tasks in the completely new end-to-end manner rather than relying on the legacy NLP pipelines. The novelty of the SUMMA project approach is that all languages covered by the project (Table 1) can be embedded in the same vectorspace by means of joint multi-task learning (Collobert et al., 2011; Dong et al., 2015; Pham, Luong & Manning, 2015) of eight LSTM-RNN translational autoencoders with hidden layer parameters shared as illustrated in Fig.2. Sharing the same vectorspace for sentences in all project languages enables accurate multilingual news story clustering without resorting to the clustering of the less accurate target (English) language machine translations. This shared vectorspace approach extends also to the unsupervised multi-task learning of language models from the large monolingual corpora (Fig. 3), which is crucial for low-resourced languages: having a generic language model learned in parallel from the monolingual corpora reduces (Dai & Le, 2015) the need for large supervised parallel corpora to achieve the same translational accuracy for the Fig. 2 setup.

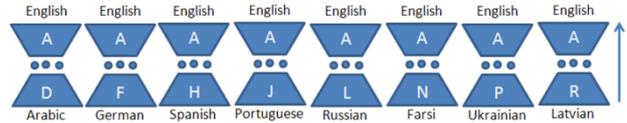

Figure 2: Supervised multi-task training of 8 translational autoencoders with shared output parameters A.

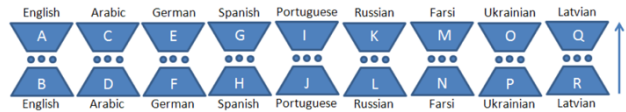

Figure 3: Unsupervised multi-task training of same-language autoencoders sharing parameters A, D, F, H, J, L, N, P, R with translational autoencoders in Figure 2.

The joint training of seventeen translational and same-language autoencoders with shared parameters (Fig. 2 and Fig. 3 together) to our knowledge has not been attempted so far. Even training of a single state-of-the-art sentence-level translational autoencoder requires days of GPU computing (Barzdins & Gosko, 2016)) in TensorFlow (Abadi et al., 2015) seq2seq model (Sutskerev, Vinyals & Le, 2014; Bahdanau, Cho & Bengio, 2014). To avoid complexities of asynchronous parallel training with shared parameter server (Dean et al., 2012), the architecture in Fig.2 and Fig. 3 instead can be trained using the alternating training approach proposed in (Luong et al., 2016), where each task is optimized for a fixed number of parameter updates (or mini-batches) before switching to the next task (which is a different language pair). Although such alternating approach prolongs the training process, it is preferred for simplicity and robustness reasons.
Once produced within SUMMA project, these translational autoencoders with shared vectorspace will be a unique language resource of likely interest also to the wider NLP community for multilingual applications outside the media monitoring domain.

## 3. Character-level Neural Translation for Streams

Neural translation attention mechanism (Bahdanau, Cho & Bengio, 2014) has been shown to be highly beneficial for bi-lingual neural translation of long sentences, but it is not compatible with the multi-task multilingual translation models (Dong et al., 2015; Luong et al, 2016) described in the previous Section and character-level translation models (Barzdins & Gosko, 2016) described in this Section. For these reasons we replace the neural translation attention mechanism with much simpler sliding-window translation approach to cope with long sentences (or potentially unsegmented streams of words produced by ASR systems during audio and video media transcription) in both multilingual and character based neural translation. Sliding-window approach quality-wise cannot compete with the state-of-art translation systems, but is adequate for fast previewing of multilingual content, especially for ASR audio transcripts with inevitably high word error rate (WER) anyway.

| The | overwhelming | Vairumu majority | Eiropas of | Savienības Union | jūras maritime | transporta traffic | pārvadā transits | caur through | | jūras maritime | ostām ports | of | | the | Eiropas European | transporta transport | tīklā network |
|---|---|---|---|---|---|---|---|---|---|---|---|---|---|---|---|---|---|
| Vairākums | apsveicamu Vairākums | Eiropas attiecībā Vairumam | Savienības uz Eiropas | jūras Eiropas Savienības jūras Savienības | Savienības jūras jūras rakstiski | transpastivē transpiltts transpirma PT rakstiski | pārraudzīts tranzītu Pārraidīt pārvietojumi Sarunas | pārvadātās tirdzniecības pārrobežu par Ar | pārvadāšana jautājumā jūras jūras pārvadātāju | izstrādāt ostu tiesību ierosinājumi rakstiski | jautājumu aktu Eiropas PT Eiropas | iespējas Eiropas transporta no | Eiropas transporta . Eiropas transporta | transporta tīkls . transporta | apliecinājuma . tīkls | . | |
| | | | | | | | | | | | | | | | | | |
| Vairākums | | Eiropas | Savienības | jūras | | transp- | | pārvadā- | | jūras | | Eiropas | transporta | tīkls | . | | |

Table 2: English-Latvian sliding window stream translation. Final translation (bottom) consists of words appearing at least twice in the neighboring columns. Word suffixes are ignored if the initial 6 characters match.

Additional problem with the low-resourced languages (bottom part in Table 1) is that most of them are highly inflective languages with rich morphology. For example, in Latvian each verb has more than 250 inflected forms. Applying regular word-level seq2seq neural translation model to such languages is impractical due to exploding vocabulary size leading to high rate of out of vocabulary (OOV) word forms and poor coordination. A better approach to translating such languages is to use the character-level translation model (Barzdins & Gosko, 2016; Karpathy, 2015; Jozefowicz, 2016). Moving from word-level to character-level neural translation makes it even harder to cope with long sentences presenting additional reason to employ the sliding-window translation approach. Table 2 illustrates the character-level neural translation from English to Latvian using modified [2] TensorFlow (Abadi et al., 2015) seq2seq (Sutskerev, Vinyals & Le, 2014) neural translation model. The character-level neural translation is enabled by forcing tokenizer to treat each input symbol as a separate "word" leading to the small and fixed "vocabulary" containing only 90 most frequently encountered characters. Another necessary change to the TensorFlow default seq2seq settings is disabling the attention (Bahdanau, Cho & Bengio, 2014) mechanism which is known to interfere with character-level translation (Barzdins & Gosko, 2016) because there are no mappings between the characters of the translated words. The small vocabulary of 90 words automatically disables also the sampled softmax functionality of seq2seq improving the overall performance. Finally, we configure single bucket of size 100 characters, which will be the max translation window size. Other hyperparameters used are: 1 LSTM layer of size 400, batch size 16. Training is performed on Europarl v7 EN-LV corpus[3] for 24h on TitanX GPU.

The sliding-window mechanism is used only during decoding (translation), mapping a fragment of 6 English words into 5 Latvian words (Latvian translations typically contain less words than English source – rich morphology substitutes for most prepositions and articles). The multiple sliding-window translations produced are later merged into the final translation consisting only of words appearing at least twice in the neighboring sliding window columns (word suffixes are ignored if the initial 6 characters of the words match – this reduces word drop due to inflection errors).

The final translation in Table 2 (bottom row) is close to the manual verbatim translation (top row) and conveys the topic of the original sentence. Moreover, the sliding-window translations are surprisingly fluent Latvian phrases with correct word forms and mostly correct coordination. The only non-Latvian "words" fabricated by the character-level translation in the Table 2 are "transpastivēšana", "transpitts", "transpirma" and apparently are triggered by the English verb "transits", because in Latvian "tranzīts" is used only as a noun without a close substitute verb.

Sliding-window translation method, obviously, cannot handle long-range dependencies well and occasionally drops or inserts words in the translation – therefore SUMMA provides also state-of-the-art translation service in parallel to the one described here. Meanwhile the sliding-window character-based translation method has unique advantages relevant to the scope of SUMMA project, discussed in the next Section.

## 4. Potential Applications of the Multilingual Character-level Stream Translation

Having a shared vectorspace multilingual translation system (Fig.2 and Fig.3) able to operate on unsegmented streams of text have a number of novel applications.

The most straightforward novel application is the possibility to embed the documents of all project languages into the same shared semantics vectorspace and compute document semantic similarity (Hill, Cho & Korhonen, 2016) irrespective of the document language. The sliding-window translation approach allows to view the document as a sequence (trace) of vectors corresponding to every sliding-window step while translating the document. These vectors are similar to word embedding vectors, but are likely to be semantically richer, as they would mostly distinguish word-senses in the context of the window. Such vector-traces corresponding to the documents can be compared in the bag-of-words fashion by measuring cosine-distance between the sums of document trace-vectors (as part of k-means clustering or nearest-neighbor search). This can be used as a building block for multilingual semantic clustering of stories into storylines, or for the semantic search of the documents in any language which are similar to the given document.

---

[2] https://github.com/didzis/tensorflowAMR

[3] http://www.statmt.org/europarl/

| English original | Latvian manual translation | Latvian character-level neural translation |
|---|---|---|
| In the last ten years, Europe has not made | Pēdējo desmit gadu laikā Eiropa nav palielināj | Pēdējo desmit gadu laikā Eiropa nav panākusi |
| We need to conclude the Bologna Process. | Mums jāpabeidz Boloņas process. | Mums jāpabeidz Boloņas process. |
| We need to promote, facilitate and open u | Mums jāveicina, jāsekmē un jānodrošina vairā | Mums jāveicina veicināt kā atvērtu un aizvien |
| We are no longer competing country again | Mēs vairs nekonkurējam valsts ar valsti, bet k | Mēs vairs nekonkurējot pret valsti, bet kā Eiro |
| If we do not make the most of the synergy | Ja mēs maksimāli neizmantosim sinerģiju, kur | Ja mēs nepadaram vislielāko daļu sinerģiju pa |
| We will be spectators, not leading players. | Mēs būsim skatītāji, nevis vadošie dalībnieki. | Mēs būsim spektori, nevis vadošie dalībnieki. |
| The way forward is the Union: more comm | Ceļš uz priekšu saistīts ar Eiropas Savienību: V | Tā ir vienotā vienotā ekonomikas politika, eko |
| Not putting up more barriers, but removing | Nevis jaunu ierobežojumu noteikšana, bet ier | Ne vairāk barjeru, bet nevis samazinājošāks, n |
| We have confidence in the Commission reg | Mēs paļaujamies uz Komisiju attiecībā uz 2020 | Mums ir uzticība Komisijā, kas attiecas uz šo 0 |
| We are convinced that the debate that is g | Mēs esam pārliecināti, ka debatēm, kas norisi | Mēs esam pārliecināti, ka debates notiks Eirop |
| Economic change and political change, and | Ekonomiskās pārmaiņas, politiskās pārmaiņas | Ekonomiskā pārmaiņas un politiskā grozījumi |
| The Treaty of Lisbon establishes new instit | Lisabonas līgums izveido jaunus amatus: pastā | Lisabonas līgums izveidojas jaunas iestāžu sta |
| It strengthens Parliament, the heart of Eur | Tas nostiprina Eiropas demokrātijas sirds - Par | Tas nostiprina Eiropas demokrātijas sirds un s |
| I can make a commitment before Parliame | Parlamentam, kas pārstāv visus Eiropas pilsoņ | Es varu padarīt parlamentu, kas pārstāv visus |
| We want those institutions to have the me | Mēs vēlamies, lai šie amati ir tādi, kā noteikts | Mēs vēlamies, lai šīs institūcijas ir tie, kas izvei |
| We are aware that this six-month period w | Mēs apzināmies, ka šis sešu mēnešu periods b | Mēs apzināmies, ka šis sešus mēneša pirmais p |
| We are going to do this, and I hope that we | Mēs to darīsim un es ceru, ka mūs šī perioda b | Mēs to darīsim, un es ceru, ka mēs tiksim piet |
| There are various powers governing the Eu | Eiropas Savienību pārvalda dažādas varas, kan | Ir dažādi pilnvaras, un tām ir jābūt kopējai tird |
| That is how we shall work. | Tā mēs strādāsim. | Tā mēs strādāsim. |
| Mr President, ladies and gentlemen, we are | Priekšsēdētāja kungs, dāmas un kungi, pārmai | Priekšsēdētāja kungs, dāmas un kungi! Mēs es |
| There will also be changes because in this c | Izmaiņas būs vērojamas arī tādēļ, ka saistībā a | Pastāv arī tādēļ, ka šī globalizācijas konteksts |

Table 3: Europarl v7 training corpus fragment (only first 100 characters of each sentence were used for training) and the character-level neural translation output illustrating the memorization of the training corpus.

Another novel application of the character-level neural translation is stream segmentation into the individual stories – a difficult task for news ingested from audio or video sources and transcribed with ASR and thus lacking any explicit sentence or story segmentation information.

For stream segmentation into the stories it is possible to utilize the exceptional generalization and memorization capacity of the neural networks, which is already applied in the neural dialogue systems such as Gmail Smart Replies (Corrado, 2015; Vinyals&Le, 2015). Table 3 illustrates how mere 400 LSTM cells of our single-layer 90-character neural translator have been able to generalize and memorize rather correct translations for the first 100 characters of the entire Europarl v7 EN-LV training corpus containing 600,000 sentence pairs.

For story segmentation a sliding-window neural translation system can be incrementally trained to monolingually "translate" the current 5 words of the stream into the next 5 words of the stream (predicting next 5 words from previous 5 words), based on the actual news streams encountered. Such system should be able to predict reasonably well the next 5 words within the news story, but will fail to do so when there is a transition from one story to the next. Along with additional auxiliary information such as time-code (when exactly the phrase was spoken and pauses in the speech) and speaker identification for each phrase this should provide a rather reliable segmentation signal.

## 5. Conclusions

It is still an open issue which vectorspace projections yield the semantically best clusters (Hill, Cho & Korhonen, 2016) and further experiments are needed. Particularly for storyline (Rissen et al., 2013) clustering the signals for the stories belonging to the same storyline might be not so much the semantic similarity of the articles (they might report various developments of the storyline from differing viewpoints), but rather the matching time and location as well as same organizations and people being involved – the information typically supplied by Named Entity Linking (NEL) tools.

The tradeoffs between semantic clustering quality and computational complexity are likely to be crucial. Once trained, the run-time use of the multilingual translation modules for translation and news story clustering is around 1 sec on TitanX GPU per average news story. This is an order of magnitude slower than regular NEL or IF IDF bag-of-words based clustering methods. Establishing reliable storyline clustering benchmarking data sets and metrics is one of the goals of the SUMMA project, as good storyline clusters are the prerequisite for downstream storyline summarization, visualization, and predictive anticipation of upcoming developments.

## 6. Acknowledgments

This work was supported in part by H2020 SUMMA project under grant agreement 688139/H2020-ICT-2015 and in part by the Latvian National research program SOPHIS under grant agreement Nr.10-4/VPP-4/11.